%% file: acl_latex.tex
\pdfoutput=1
\documentclass[11pt]{article}

\usepackage{acl}

\usepackage{times}
\usepackage{latexsym}

\usepackage[T1]{fontenc}
\usepackage[utf8]{inputenc}

\usepackage{microtype}

\usepackage{inconsolata}

\usepackage{graphicx}
\usepackage{lipsum}
\usepackage{algorithm}
\usepackage[noEnd=false]{algpseudocodex}
\usepackage{tablefootnote}
\usepackage{threeparttable}

\title{BAMBINO-LM: (Bilingual-)Human-Inspired Continual Pre-training of BabyLM}

\author{Zhewen Shen$^1$ Aditya Joshi $^1$ Ruey-Cheng Chen$^2$ \\
  $^1$ University of New South Wales, Sydney, Australia \\
  $^2$ Canva, Sydney, Australia\\
  \texttt{zhewen.shen@student.unsw.edu.au}, \texttt{aditya.joshi@unsw.edu.au}, \texttt{rcchen@canva.com} \\
}

\usepackage{booktabs}
\usepackage{array}
\usepackage{amsfonts, amsmath, amssymb, amsthm}
\usepackage{footnote}
\makesavenoteenv{tabular}

\begin{document}
\maketitle
\begin{abstract}
% Bilingual children often use feedback from a parent or teacher when learning a second language. With this inspiration, this paper investigates how bilingual ability can be imparted to small-scale language models (specifically, BabyLM).
Children from bilingual backgrounds benefit from interactions with parents and teachers to re-acquire their heritage language. In this paper, we investigate how this insight from behavioral study can be incorporated into the learning of small-scale language models. We introduce BAMBINO-LM, a continual pre-training strategy for BabyLM that uses a novel combination of alternation and PPO-based perplexity reward induced from a parent Italian model. Upon evaluation on zero-shot classification tasks for English and Italian, BAMBINO-LM improves the Italian language capability of a BabyLM baseline. Our ablation analysis demonstrates that employing both the alternation strategy and PPO-based modeling is key to this effectiveness gain. We also show that, as a side effect, the proposed method leads to a similar degradation in L1 effectiveness as human children would have had in an equivalent learning scenario. Through its modeling and findings, BAMBINO-LM makes a focused contribution to the pre-training of small-scale language models by first developing a human-inspired strategy for pre-training and then showing that it results in behaviours similar to that of humans. 
\end{abstract}

\section{Introduction}
\input{sections/introduction}

\section{Related Work}
\input{sections/related-work}

\section{Methods}
\input{sections/methods}

\section{Experiment Setup}
\input{sections/expsetup}
\section{Results}

\input{sections/results}

\section{Conclusion}
This paper introduces BAMBINO-LM, a continual pre-training strategy mimicking the process of second language acquisition in an interactive setting. BAMBINO-LM uses a two-phase approach: it incorporates reward from a parent Italian model into a PPO-based mechanism and alternates this procedure together with causal language modeling based on Italian language text. Our experiments demonstrate systematic improvement in Italian with a marginal but expected decrease in English, which echoes the past results in second language acquisition for large language models~\cite{evanson2023language}. These findings highlight the efficacy of our approach in enhancing bilingual capabilities while maintaining performance in the original language.

In future work, we aim to explore the effect of alternative metrics and different reward learning mechanisms that better align with human feedback behaviors. This also includes exploring rewards that capture linguistic quality and provide direct, ``constructive'' corrections to the model output which is commonly known as an effective learning strategy for language development. Although BAMBINO-LM was applied for second language learning with Italian as an example, the method must be validated for other languages, especially languages that are distant from English or those that use a different set of tokens. The degradation in the performance of the first language, English, points to the potential of alternating with language modeling for the first language.

\section*{Limitations}
The approach relies on the availability of a base model in a language, English in our case. Although we download Italian language datasets from known Italian sources, we do not explicitly validate the language of the text. We use the PPO model as is, and do not experimentally tune its parameters. Similarly, using perplexity as a metric for computing rewards may not be the optimal solution, as perplexity itself is influenced by many factors of the parent model. 

\section*{Ethics Statement}
The paper uses publicly available datasets for training and evaluation that do not possess known harms. The evaluative tasks are typical language learning tasks. However, the resultant models are not tested for harmful or biased content.

\section*{Acknowledgment}
Zhewen Shen conducted this research as a part of UNSW Sydney's Taste of Research Program.
\bibliography{anthology,custom}

% \appendix

% \section{Example Appendix}
% \label{sec:appendix}

% This is an appendix.

\end{document}

%% file: sections/introduction.tex
The recently held \textbf{BabyLM} challenge \citep{warstadt-etal-2023-findings} explores pretraining of language models using a constrained dataset analogous to the linguistic exposure of a 13-year-old English-speaking child. In this paper, we extend the BabyLM challenge to a bilingual setting, drawing inspiration from parent-child interactions in heritage language acquisition~\cite{lohndal2019heritage}. Immigrant children in western societies, who may have acquired their home language at a young age, can sometimes need to re-acquire the same language during the school years when the language becomes a minority. These heritage speakers typically benefit from an extended exposure to the minority language at home or in the community, owing largely to feedback and stimuli provided by parents and family members~\cite{montrul2010current}.
This observation about child bilingualism is in line with the behaviorist theory for child language development~\cite{demirezen1988behaviorist}.
% Children in western societies who learn English in school and another language at home often rely on feedback from their parents for the latter.

Inspired by this line of work, we ask the following research question in the context of computational language modeling:
\begin{quote}
\emph{Can a small-scale language model trained on the majority language (e.g. English) be continually pre-trained on the minority language, leveraging the feedback of a second model that is fluent in the latter language?}
\end{quote}
To address this question, we introduce `\textbf{Bilingual language Acquisition Modeling Based on INterleaved Optimization of Language Models (BAMBINO-LM)}', a novel continual pretraining strategy that uses a combination of alternation and proximal policy optimization (PPO) using a reward from a second model playing the parent role (i.e., a large language model pre-trained in the minority language). We experiment with BabyLM trained on \textit{English}, and continually pretrain this model on an assumed second language, \textit{Italian}.In its connection to cognitive processing, our work makes the following contributions:
\begin{itemize}
    \setlength{\itemsep}{0cm}
    \item  BAMBINO-LM draws inspiration from bilingual language acquisition and learns from interactions with a second model by \textbf{incorporating a perplexity-based reward for language model pre-training}. To the best of our knowledge, this is the first work to use PPO-based modeling for language acquisition in BabyLM.
    \item We show that BAMBINO-LM can acquire Italian to a reasonable degree with some expected degradation in its English capability. The findings hint at \textbf{a common learning trajectories for second language acquisition shared by language models and humans}.
\end{itemize}

%During pre-training, BAMBINO-LM learns to incorporate feedback from an Italian language model via reinforcement learning. Our results on analogous sentiment classification tasks in English and Italian show that BAMBINO-LM performs improves its ability on Italian language tasks (UINAUIL) while suffering some degradation on English language tasks (GLUE) - again typical of bilingual children. 

%While this approach is viable for well-resourced languages like English, it presents a significant bottleneck for low-resource languages \citep{joshi-etal-2020-state}, where such data and computational assets are often scarce. 

% These approaches span techniques such as efficient sample selection (where samples are selected based on certain criteria), curriculum learning (where samples are batched appropriately), and so on. 

%% file: sections/related-work.tex
Pre-training small-scale language models is an emerging field that has garnered some interest from the language acquisition community. BabyBERTa~\cite{huebner2021babyberta} is an early adaptation to this scenario. \citet{warstadt-etal-2023-findings} introduce the BabyLM challenge to provide an atypically small dataset for benchmarking small-scale language models.
This shared task enables research in not only language acquisition but also sample-efficient pre-training. In the case of our paper, we do not focus on sample efficiency but instead describe ways to enhance the ability of a second language via continual pre-training.
% They state that their goal is to bring together researchers in sample-efficient pre-training and human language acquisition. 

Our work is conducted in a setup similar to \citet{yadavalli-etal-2023-slabert}, where a tiered first/second language acquisition process is attempted. \citet{samuel2023mean} also experiments with a teacher-student setting but only tests the approach on English tasks. \citet{evanson2023language} is another closely related work, which investigates the learning trajectory of large-scale language models by probing their syntactic and semantic capabilities at each step.
% Contrasting GPT-2 models with children, the study shows that linguistic skills are acquired sequentially in a similar order.

Conventionally in language generation for natural language processing, the design of feedback signals is commonly discussed in the context of knowledge distillation \cite{calderon-etal-2023-systematic}.
Recently, reinforcement learning from human feedback (RLHF) utilizes human preferences for serving reward signals when dealing with sparse training labels \cite{christiano2017deep, stiennon2020learning}, and has been shown successful for generative tasks such as dialogues and summarization. This approach is further extended in \citet{bai2022constitutional} by using AI feedback (RLAIF) to remove the dependency on human preference data, leading to better scalability and signal availability.
The approach we take in this paper mostly falls within the latter camp, but generally departs from all prior efforts in the way the parent model's perplexity is used to signal the conformity of the child model's generation. This is in contrast to sequence-level knowledge distillation \cite{kim2016sequence} where teacher's generation is used to guide the learning process.

%% file: sections/methods.tex
\begin{figure*}
    \centering
    \includegraphics[width=\textwidth]{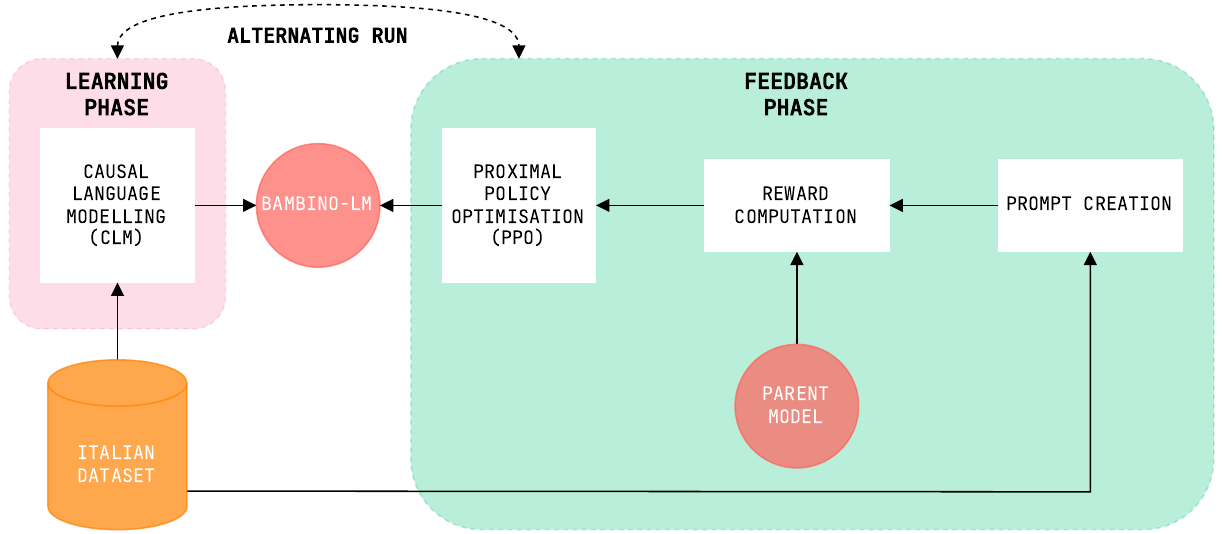}
    \caption{Architecture of BAMBINO-LM.}
    \label{fig:arch}
\end{figure*}

Figure~\ref{fig:arch} shows the two phases of BAMBINO-LM. The learning phase involves continual pre-training a small-scale language model (\emph{baby model} $\mathcal{B}$) whose initial pre-training was originally done on English data, while the feedback phase involves interactions with the Italian language model (\emph{parent model} $\mathcal{P}$). During the learning phase, pre-training for $\mathcal{B}$ is continued by employing causal language modeling on Italian data. Causal language modeling (CLM), also known as next token prediction, is a standard technique to train a decoder-only model. The objective is defined as follows:
\[
    \mathcal{L}_{\operatorname{CLM}} = -\frac{1}{|x|}\sum_{i=0}^{|x|}\log\mathbb{P} (x_t \mid x_0, \ldots, x_{t-1}).
\]
There are two architectural innovations in BAMBINO-LM:

\paragraph*{Feedback phase based on PPO}

We construct prompt $x$ by selecting the first $k$ tokens from the training example and solicit output $y_{\mathcal{B}} = \mathcal{B}(x)$ from the baby model. We then use Proximal Policy Optimization (PPO) where $\mathcal{B}$'s parameters are updated according to a clipped surrogate objective~\cite{schulman2017proximal}. This objective moderates the updates to the policy, facilitating stable and efficient learning by incorporating a clipping mechanism.
% new edit
% \begin{small}
% \[
%     \mathcal{L}_{\text{PPO}} = \hat{\mathbb{E}}_t \left[ \min \left( r_t(\theta) \hat{A}_t, \operatorname{clip}(r_t(\theta), 1 - \epsilon, 1 + \epsilon) \hat{A}_t \right) \right],
% \]
% \end{small}
Its definition is given as follows:
\[
    \mathcal{L}_{\text{PPO}} = \hat{\mathbb{E}}_t \left[ \min \left( r_t(\theta) \hat{A}_t, \operatorname{clip}(r_t(\theta); \epsilon) \hat{A}_t \right) \right],
\]
with $\theta$ being the model parameters and $r_t(\theta)$ defined as:
\[
    r_t(\theta) = \frac{\pi_\theta(a_t | s_t)}{\pi_{\theta_{\text{old}}}(a_t | s_t)}.
\]
In the autoregressive setting of language modeling, $\theta$ controls the generation of tokens based on the given state or context $s_t$. The probability ratio $r_t(\theta)$ quantifies the change in the likelihood of selecting action $a_t$ (the next token), under the updated policy parameters compared to the previous parameters $\theta_{\text{old}}$. This ratio provides understanding on the impact of parameter updates on the policy's behavior, ensuring that changes do not excessively deviate from the previous policy, thereby maintaining training stability.
The clipping mechanism, defined by $\operatorname{clip}(r_t(\theta); \epsilon)$, restricts $r_t(\theta)$ within the bound $[1 - \epsilon, 1 - \epsilon]$, mitigating the risk of large policy updates that could lead to divergence.

The advantage function, $\hat{A}_t = R + \gamma V(s_{t+1}) - V(s_t)$, which reflects the relative gains of selecting $a_t$ given $s_t$. This function guides the optimization process by favoring actions that lead to better than expected outcomes.

The reward $R$ for the advantage function is then calculated using the following function:
\begin{equation}
R(y_{\mathcal{B}}) = \frac{\alpha}{\beta(\operatorname{PPL}_{\mathcal{P}}(y_{\mathcal{B}}) - \tau)},
\end{equation}
where $\alpha$ and $\beta$ are parameters, $\operatorname{PPL}_{\mathcal{P}}$ represents the perplexity of the parent model $\mathcal{P}$ for the sequence $y_{\mathcal{B}}$, and $\tau$ is a threshold value for perplexity. We use the following formulation of perplexity:
\[
    \operatorname{PPL}(x) = \exp\left[\sum_{i = 0}^{|x|} \log \mathbb{P} (x_t \mid x_0, \ldots, x_{t-1})\right].
\]

\paragraph*{Alternating run}

We adopt an alternating run strategy between the learning and feedback phases, which is summarized in Algorithm~\ref{alg:train}. The rationale behind this is two-fold: 1) this strategy simulates frequent interactions between a child and its parent through dialogues, which has been our main motivation behind this study; 2) using multiple rewards is shown beneficial for reinforcement learning~\cite{dann2023reinforcement}. To expand on the second point, our findings further suggest that using perplexity as a reward can lead to exploitation when baby model $\mathcal{B}$ attempts to produce similar utterances to those coming from parent model $\mathcal{P}$. Without this strategic alternation between CLM and PPO, the pre-training tends to produce undesirable behaviours such as repeating words.

\begin{algorithm}
    \caption{BAMBINO-LM Training.}
    \label{alg:train}
    \begin{algorithmic}[1]
        \Procedure{Train}{$\mathcal{D}, \mathcal{B}, \mathcal{P}$}
            \State \textbf{Input:} pre-training dataset $\mathcal{D}$, baby model $\mathcal{B}$, and parent model $\mathcal{P}$.
            
            \State $r_{\operatorname{CLM}}, r_{\operatorname{PPO}} \gets 10, 2$
            \State $r \gets r_{\operatorname{CLM}} + r_{\operatorname{PPO}}$
            \For{$i, x \in \operatorname{enumerate}(\mathcal{D}$)}

            \If{$i \% r < r_{\operatorname{CLM}}$}
                \State \textbf{perform} CLM step
            \Else
                % \State $x \gets x[1..k]$
                \State $y_\mathcal{B} \gets \mathcal{B}(x[1..k])$
                \State reward $\gets R(y_\mathcal{B})$
                \State \textbf{perform} PPO step
            \EndIf

            \EndFor
        \EndProcedure
    \end{algorithmic}
\end{algorithm}

%% file: sections/expsetup.tex
Mimicking their process to create the BabyLM challenge corpus \citep{warstadt-etal-2023-findings}, we create an Italian dataset that is comparable in size to the \emph{strict-small} track of the challenge, and perform identical preprocessing\footnote{\url{https://github.com/babylm/babylm_data_preprocessing}; Accessed on 13th May, 2024.}. Table~\ref{tab:datasets} shows the statistics of the Italian language dataset. A cursory quality check was conducted to ensure that the dataset was in a readable format.

For the choice of the baby model, we use English baseline \texttt{OPT-125m}~\cite{zhang2022opt} model for the \emph{strict-small} track provided by the BabyLM organizers. For the parent model, we use \texttt{gpt2-small-italian} model by \citet{de-vries-nissim-2021-good}. Using the Italian dataset described above, we conduct continual pretraining over 10 epochs, consisting of 10 learning phase steps followed by 2 feedback phase steps. We use $k = 5$ to solicit the first few tokens for prompting the baby model. All models are trained using HuggingFace's \texttt{transformer} \citep{wolf-etal-2020-transformers} and \texttt{trl} \citep{vonwerra2022trl} library.

\input{sections/dataset-table}

For downstream tasks, we use four Italian language tasks in UINAUIL~\cite{basile-etal-2023-uinauil} and four English language tasks in GLUE~\cite{wang2018glue}. The tasks were selected primarily based on computational constraints for the project. We also include BLiMP~\cite{warstadt2020blimp} for that it is used in the original BabyLM challenge. All tasks are conducted in a \textit{zero-shot classification} setting.

%% file: sections/dataset-table.tex
\begin{table}[!ht]
    \centering
    \begin{tabular}{@{}p{0.8\columnwidth} r@{}}
        \toprule
            \textbf{Dataset} & \textbf{\%} \\
        \midrule
            CHILDES \citep{macwhinney2000childes} & 2.23 \\
            DailyDialog \citep{li-etal-2017-dailydialog} & 4.45 \\
            QED \citep{abdelali-etal-2014-amara} & 11.86 \\
            OpenSubtitles \cite{lison-tiedemann-2016-opensubtitles2016} & 27.58 \\
            Standardised Project Gutenberg Corpus \cite{e22010126} & 16.19 \\
            Children's Story \tablefootnote{\url{https://www.gutenberg.org/ebooks/bookshelf/353}} & 18.57 \\
            Wikipedia \tablefootnote{\url{https://dumps.wikimedia.org/itwiki/}} & 19.10 \\
        \bottomrule
    \end{tabular}
    \caption{Italian dataset used for continual pre-training.}
    \label{tab:datasets}
\end{table}

%% file: sections/results.tex
\input{sections/results-table-main}

Table~\ref{tab:result} shows a significant improvement in Italian downstream tasks for BAMBINO-LM as compared with the BabyLM baseline. Specifically, we achieve an average improvement of 0.1197 $(0.3416 \to 0.4613)$ without substantial differences in English classification tasks. However, we notice an expected decrease of 0.0752 $(0.6255 \to 0.5503)$ in the English language BLiMP dataset. These observations are in line with \citet{yadavalli-etal-2023-slabert} which show that native child-directed speech can lead to negative cross-lingual transfer and impede L2 acquisition depending on the choice of L1.

% BAMBINO-LM combines two novel ideas: alternating runs and the use of PPO.
In Table~\ref{tab:ablation} we examine two ablated versions of our model: (a) \textbf{w/o PPO:} Trained solely on the CLM objective with no feedback phase; (b) \textbf{w/o alternating:} Use BAMBINO-LM with no alternating runs. Instead, it trains with the CLM objective for the first $85\%$ of each epoch and then switches to the PPO objective for the remaining $15\%$.

Removing the interactive feedback mechanism (w/o PPO) and the alternating strategy (w/o alternating) significantly decreases Italian performance compared to our primary model. On UINAUIL tasks, the average score drops from 0.4613 to 0.4000 (w/o PPO) and 0.3513 (w/o alternating). However, we do not observe significant improvements in performance in both the English language task sets (GLUE and BLiMP). For GLUE tasks, the average scores remain consistent, with 0.4373 for BAMBINO-LM, 0.4375 for w/o PPO, and 0.4357 for w/o alternating. On the BLiMP dataset, the average scores are 0.5503 for BAMBINO-LM and 0.5554 for w/o PPO.

\input{sections/results-table-abl}

These results indicate that PPO modeling and alternating runs are both crucial for improving the bilingual ability of BAMBINO-LM without negatively impacting English performance. Furthermore, the lack of significant changes in English scores reinforces that these strategies enhance bilingual capabilities without compromising performance on existing benchmarks.

%% file: sections/results-table-main.tex
\begin{table}[ht]
\centering
\begin{tabular}{lcc}
\toprule
\toprule
\textbf{Task / Model} & BabyLM &  BAMBINO-LM \\
\midrule
\multicolumn{3}{c}{\textbf{UINAUIL}} \\
\midrule
HaSpeeDe               & \textbf{0.4774}     & 0.4592 \\
IronITA                & 0.4966     & \textbf{0.5516} \\
SENTIPOLC              & 0.1575     & \textbf{0.4050} \\
Textual Entailment     & 0.4950     & \textbf{0.5525} \\
\textit{Average}       & 0.3416     & \textbf{0.4613} \\
\midrule
\multicolumn{3}{c}{\textbf{GLUE}} \\
\midrule
MNLI                   & 0.3472           & \textbf{0.3530} \\
MNLI-MM                & 0.3483           & \textbf{0.3521} \\
RTE                    & \textbf{0.5271}           & 0.5199 \\
SST2                   & 0.5034           & \textbf{0.5241} \\
\textit{Average}       & 0.4315           & \textbf{0.4373} \\
\midrule
\multicolumn{3}{c}{\textbf{BLiMP}} \\
\midrule
\textit{Average}      & \textbf{0.6255} & 0.5503 \\
\bottomrule
\bottomrule
\end{tabular}
\caption{Comparison of BAMBINO-LM with BabyLM.}
\label{tab:result}
\end{table}

%% file: sections/results-table-abl.tex
\begin{table}[!ht]
\centering
\begin{tabular}{p{2cm}p{2.3cm}p{2.3cm}}
\toprule
\toprule
\textbf{Task / Model} & BAMBINO-LM w/o PPO & BAMBINO-LM w/o alternating \\
\midrule
\multicolumn{3}{c}{\textbf{UINAUIL}} \\
\midrule
HaSpeeDe               & 0.4798   & 0.4925 \\
IronITA                & 0.4966   & 0.4989 \\
SENTIPOLC              & 0.2775   & 0.1580 \\
Textual Entailment     & 0.5500   & 0.5500 \\
\textit{Average}       & 0.4000   & 0.3513 \\
\midrule
\multicolumn{3}{c}{\textbf{GLUE}} \\
\midrule
MNLI                   & 0.3540   & 0.3522  \\
MNLI-MM                & 0.3502   & 0.3545  \\
RTE                    & 0.5343   & 0.5271  \\
SST2                   & 0.5115   & 0.5092 \\
\textit{Average}       & 0.4375   & 0.4357\\
\midrule
\multicolumn{3}{c}{\textbf{BLiMP}} \\
\midrule
\textit{Average}      & 0.5554   & 0.5268\\
\bottomrule
\bottomrule
\end{tabular}
\caption{Results of the ablation experiments.}
\label{tab:ablation}
\end{table}